\documentclass[arxiv]{melba}
\usepackage{mwe} % to get dummy images
\usepackage{amsfonts}
\usepackage{hyperref}
\usepackage{soul}

 \melbaauthors{}  % Note: this one is also used to set the pdf 'authors' metadata
\melbayear{2025}  % The publication year
\melbaspecialissue{Medical Imaging with Deep Learning (MIDL) 2025}
\ShortHeadings{Network-induced shape artifact detection}{Network-induced shape artifact detection}

\title{Knowledge-based anomaly detection for identifying network-induced shape artifacts}
\author{
    \begin{center}
    \hspace{0.12cm}
         \name Rucha Deshpande,
         \name Tahsin Rahman,
        \name Miguel Lago,
        \name Adarsh Subbaswamy,
        \name Jana G. Delfino,
        \name Ghada Zamzmi, 
        \name Elim Thompson, 
        \name Aldo Badano, 
        \name Seyed Kahaki
    \hspace*{0.12cm}
    \end{center}
}

\affiliations{% <- trailing '%' to avoid unwanted indent
    \begin{center}
    \vspace{-0.9cm}
	Division of Imaging, Diagnostics, and Software Reliability, Office of Science and Engineering Laboratories, Center for Devices and Radiological Health, U. S. Food and Drug Administration, USA \\
    \end{center}
}

\email{Seyed.Kahaki@fda.hhs.gov}

\abstract{%   <- trailing '%' for backward compatibility of .sty file
	Synthetic data provides a promising approach to address data scarcity for training machine learning models; however, adoption without proper quality assessments may introduce artifacts, distortions, and unrealistic features that compromise model performance and clinical utility.
    This work introduces a novel knowledge-based anomaly detection method for detecting network-induced shape artifacts in synthetic images. 
    The introduced method utilizes a two-stage framework comprising (i) a novel feature extractor that constructs a specialized feature space by analyzing the per-image distribution of angle gradients along anatomical boundaries, and (ii) an isolation forest-based anomaly detector. 
    This representation captures local shape variations while maintaining global anatomical correspondence regardless of size differences. The isolation forest is trained on patient data to learn normal anatomical shape characteristics and assigns anomaly scores to synthetic images.
    This method can be used to detect anatomically unrealistic images irrespective of the generative model used and provides interpretability through its knowledge-based design.
    We demonstrate the effectiveness of the method for identifying network-induced shape artifacts in two synthetic mammography datasets generated by different architectures: a latent diffusion model and StyleGAN2, trained on CSAW-M and VinDr-Mammo patient datasets respectively. Quantitative evaluation shows that the method successfully concentrates artifacts in the most anomalous partition (1st percentile), with AUC values of 0.97 (CSAW-syn) and 0.91 (VMLO-syn). In addition, a reader study involving three imaging scientists confirmed that images identified by the method as containing network-induced shape artifacts were also flagged by human readers with mean agreement rates of 66\% (CSAW-syn) and 68\% (VMLO-syn) for the most anomalous partition, approximately 1.5-2 times higher than the least anomalous partition. Kendall-Tau correlations between algorithmic and human rankings were 0.45 and 0.43 for the two datasets, indicating reasonable agreement despite the challenging nature of subtle artifact detection. 
    This method is a step forward in the responsible use of synthetic data, as it allows developers to evaluate synthetic images for known anatomic constraints and pinpoint and address specific issues to improve the overall quality of a synthetic dataset.
}


\begin{document}

% top matter
\twocolumn[\maketitle]
% comment the preceedings and uncomment the following if the authors list + abstract is longer than one page
% \maketitle
% \twocolumn

%%%%%%%%%%%%%%%%%%%%%%%%%%%%%%%%%%%%%%%%%%%%%%%%%%%%%%%%%%%%%%%%%%%%%%%
% Introducion
%%%%%%%%%%%%%%%%%%%%%%%%%%%%%%%%%%%%%%%%%%%%%%%%%%%%%%%%%%%%%%%%%%%%%%%
% \rule{\textwidth}{1pt}
\section{Introduction}
Deep learning has improved medical imaging, particularly for automated image analysis applications and clinical decision-making. However, the development of robust deep learning models is hindered by limited access to large-scale, high-quality patient datasets. This limitation is further worsened by issues such as patient privacy concerns~\citep{Giuffre2023-wv}, unavailability of data for rare diseases~\citep{Vrudhula2024-gg}, and low quality (or even the lack) of truth labels. The limited availability of high-quality datasets restricts the ability to train and validate deep learning models, leading to over-fitting, poor generalization, and reduced accuracy when applying the models to unseen real-world data~\citep{Yang2024-iy}.

%The use of synthetic data has emerged as a promising solution for augmenting patient datasets with known truth labels while safeguarding patient privacy.
Synthetic data, defined as ``artificial data that is intended to mimic the properties and relationships seen in real patient data"~\citep{fda_aiml_glossary_2024}, offers a promising solution to address data-related limitations in healthcare~\citep{Sizikova2024-gk}. By generating synthetic data with known truth labels, the costs associated with data collection and acquisition can be significantly reduced. Furthermore, synthetic data can be used to augment real patient datasets, particularly for patients with rare diseases and/or from underrepresented populations, thereby reducing dataset bias due to class imbalance and potentially improving performance generalizability of deep learning algorithms in medical imaging~\citep{Chen2021-ez, Garcea2023-xq}. 

Various techniques have been developed to generate synthetic data, ranging from knowledge-based approaches \citep{badano2018evaluation,kim2024s} to advanced generative artificial intelligence (AI) methods \citep{kazerouni2023diffusion,showrov2024generative}. Based on physical modeling, knowledge-based approaches preserve physical constraints between attributes and physical findings~\citep{Sizikova2023-ab}, whereas generative AI approaches learn from large datasets and capture complex patterns and relationships in the data~\citep{Jeong2022-oe,Koetzier2024-kp}. Despite these advancements, assessing the quality and clinical relevance of synthetic medical images, particularly those generated by generative AI models, remains a challenge \citep{deshpande2025report, muller2023multimodal}. 

\begin{figure*}[t]
    \centering
    \includegraphics[width=1\linewidth]{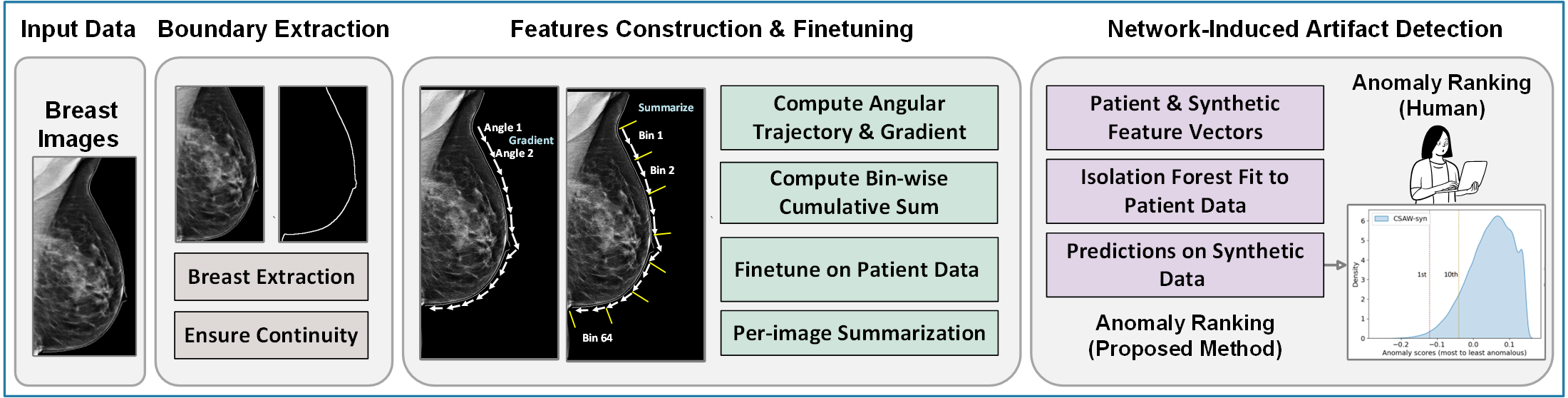}
    \caption{Overview of the proposed method for detecting network-induced shape artifacts.
    }
    \label{fig:overview}
\end{figure*}

A key challenge with synthetic data obtained from generative AI models is the occurrence of unrealistic features or artifacts, which can arise when models prioritize matching overall data distributions over preserving fine-grained, image-level details or adherence to clinical or anatomical constraints. 
Figure~\ref{fig:synExamples} shows examples of synthetic mammography images from datasets created by training different generative AI models on the CSAW-M ~\citep{Pinaya2023-kd} and VinDr-Mammo \citep{pham2022vindr} patient datasets respectively. The red annotations in these images highlight various AI-generated artifacts, including unnatural geometric patterns, synthetic noise within breast tissue, and artificial discontinuities along tissue boundaries. These artifacts demonstrate failure modes where the generative AI model inadequately capture the natural anatomical shapes and morphological structures in real patient mammography, potentially introducing systematic biases that could compromise the diagnostic accuracy and clinical reliability of downstream AI models trained for lesion detection and classification tasks.

While network-induced shape artifacts in synthetic images are well-documented in the literature \citep{lee2023impact,muller2023multimodal,deshpande2025report,kelkar2023assessing,schwarz2021frequency}, automated methods for identifying such artifacts in individual images remain scarce \citep{deshpande2025report}. Popular evaluation methods typically rely on dataset-wide metrics \citep{borji2022pros,Zamzmi2025-mf}, which summarize overall distribution alignment in a feature space. While useful for assessing general trends, these metrics overlook localized artifacts that may appear only in a fraction of the dataset, thus making it difficult to identify individual distorted images. This lack of granular assessment is particularly problematic because synthetic datasets often contain a vast number of images, which makes visual assessment of artifacts in individual images challenging.

While there is a potential need to evaluate the quality of individual synthetic images, such assessments in large synthetic datasets have unique challenges: (i) low-prevalence artifacts may be missed in visual spot-checking, which fails to provide a comprehensive evaluation of all images in the dataset, (ii) the types of artifacts induced by networks are often unknown and artifact labels typically unavailable, (iii) artifacts may vary across different generative models, and (iv) domain-specific factors, such as anatomy or imaging protocols, add further complexity to automated artifact detection. Thus, there is a need for domain-relevant methods that assess individual images for the presence of network-induced shape artifacts in synthetic datasets. 

In this paper, we propose a method, outlined in Figure~\ref{fig:overview}, for detecting network-induced shape artifacts using a knowledge-based feature space that captures the shape characteristics of the anatomy of interest. This feature space is constructed by analyzing the per-image distribution of angle gradients along the boundary of the anatomical region of interest. Building on this representation, artifact detection is performed using an isolation forest \citep{liu2008isolation}. The isolation forest is trained on a patient dataset to capture the shape characteristics of real data and is then applied to the corresponding synthetic dataset. Each synthetic image is assigned an anomaly score, where highly negative scores indicate a higher likelihood of containing artifacts, while non-negative and high positive scores correspond to normal images.
The proposed method (i) can identify images with unrealistic anatomical shapes, (ii) can greatly improve the efficiency of visual search by ranking a dataset of images, (iii) is model-agnostic due to the features being obtained from the generated images alone, and (iv) is interpretable due to its knowledge-based design. This method allows each image in a synthetic dataset to be evaluated for adherence to known anatomical constraints. As a result, developers can pinpoint and address specific issues and improve the overall quality of synthetic datasets in a targeted and efficient manner.

%%%%%%%%%%%%%%%%%%%%%%%%%%%%%%%%%%%%%%%%%%%%%%%%%%%%%%%%%%%%%%%%%%%%%%%
% Materials & Methods
%%%%%%%%%%%%%%%%%%%%%%%%%%%%%%%%%%%%%%%%%%%%%%%%%%%%%%%%%%%%%%%%%%%%%%%

\section{Materials \& Methods}
\subsection{Datasets}
While our proposed method is model and anatomy agnostic, we demonstrate its utility using two synthetic digital mammography datasets in the mediolateral oblique (MLO) view. Each synthetic dataset was generated via a different generative model. 

\subsubsection{SinKove}

The first synthetic dataset, SinKove, is a public dataset \citep{pinaya2023generative} with 100,000 images generated by a latent diffusion model trained on the CSAW-M patient dataset \citep{sorkhei2021csaw}. The latent diffusion model comprises a variational autoencoder with KL-regularization to learn the latent representation from the patient data and then a diffusion U-Net to learn a generative model of the latents.

CSAW-M \citep{sorkhei2021csaw} is a publicly available mammography dataset which consists of about 10,000 images designed for non-commercial use, hosted by the SciLifeLab Data Repository, a Swedish infrastructure for sharing life science data. It provides screening mammograms accompanied by metadata, expert annotation of lesions,  clinical endpoints, density measures, and image acquisition parameters. The dataset comprises a training set (9,523 examples), a public test set (497 examples), and a private test set (475 examples) derived from the CSAW cohort, a collection of millions of mammograms from screening participants aged 40–74 collected between 2008 and 2015.

Images in CSAW-M were curated from participants at Karolinska University Hospital, using the most recent mediolateral oblique (MLO) view for optimal breast visualization. To avoid contamination from tumor presence, contralateral breast images were used for cancer cases, while random breast sides were selected for non-cancer cases. Images were preprocessed to ensure uniformity, including resizing, intensity scaling, zero-padding, and removal of text annotations. This resulted in 632×512, 8-bit PNG images suitable for analysis. Examples of CSAW-M images are shown in Figure~\ref{fig:image_samples}.

These real and synthetic datasets are hereafter referred to as CSAW-real and CSAW-syn, respectively.

\subsubsection{VMLO-syn}

The second synthetic dataset, VMLO-syn, was generated using StyleGAN2 \citep{karras2020analyzing} trained on the MLO view subset (VMLO-real) of VinDr-Mammo dataset \citep{pham2022vindr}, which contains approximately 10,000 images. We trained StyleGAN2 on Nvidia A100 GPUs. 

\begin{figure}[t]
    \centering
    \includegraphics[width=\linewidth]{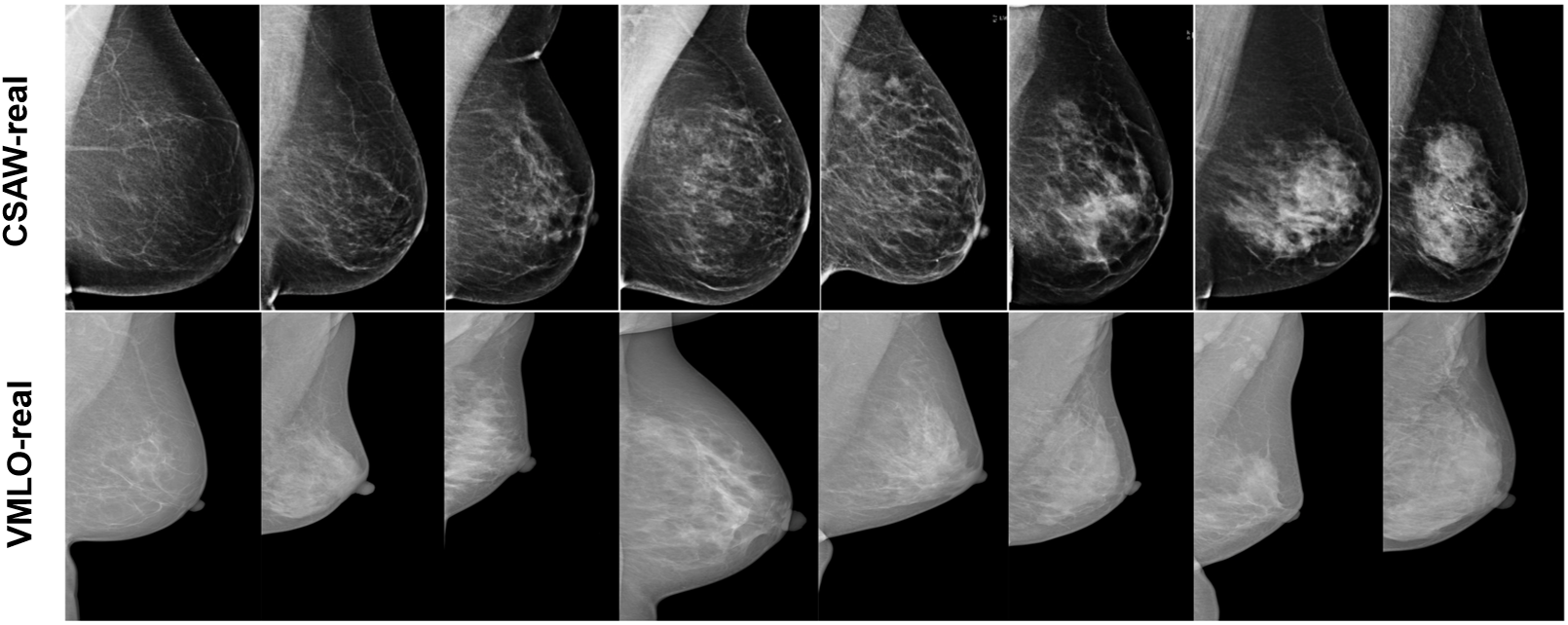}
    \\[0.3em]
    {\small (a) Patient data samples}
    \\[1em]
    \includegraphics[width=\linewidth]{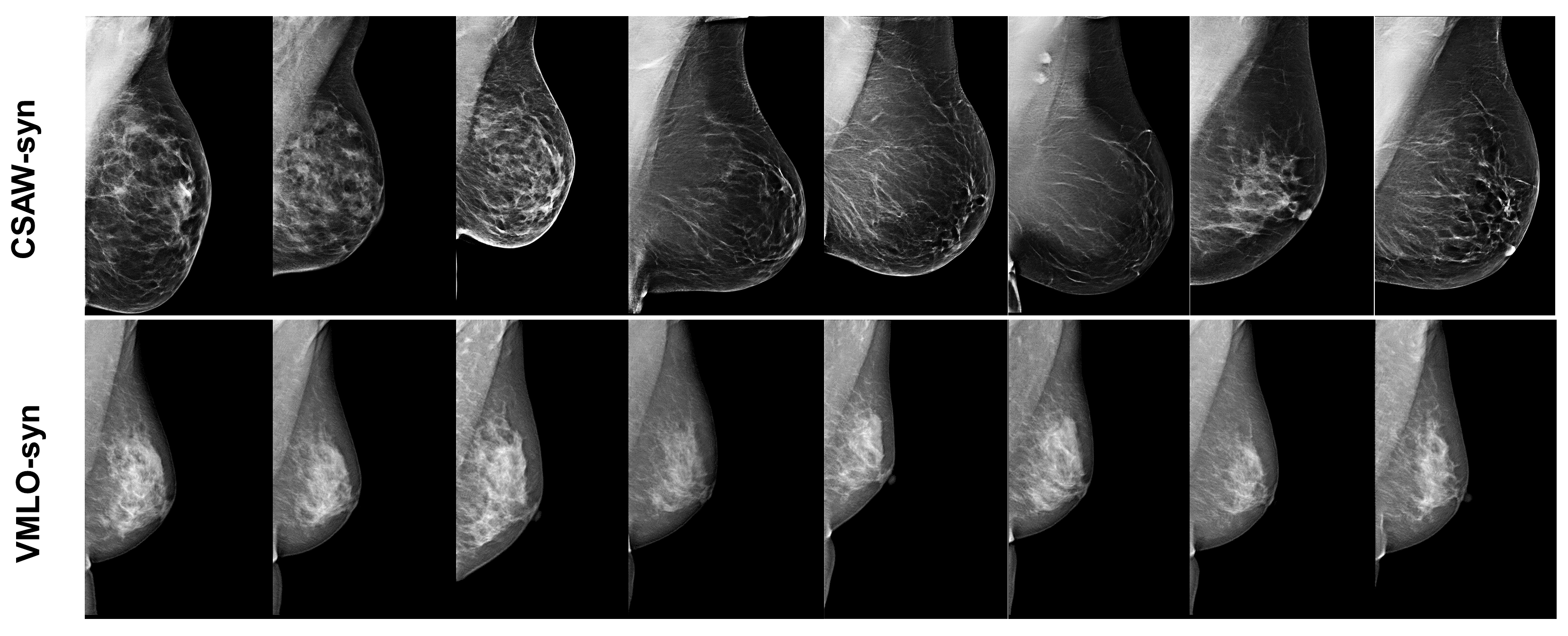}
    \\[0.3em]
    {\small (b) Synthetic data samples}
    \caption{Examples of images from (a) patient datasets: CSAW-real (upper) and VMLO-real (lower) and (b) synthetic datasets: CSAW-syn (upper) and VMLO-syn (lower).}
    \label{fig:image_samples}
\end{figure}

The VinDr-Mammo dataset \citep{pham2022vindr} is a publicly available, comprehensive collection of mammography images created to advance research in computer-aided detection (CADe) and diagnosis (CADx) systems for breast cancer screening. The dataset consists of 20,000 mammography images in DICOM format, sourced from 5,000 exams conducted between 2018 and 2020 at Hanoi Medical University Hospital (HMUH) and Hospital 108 (H108) in Vietnam. The VinDR-Mammo dataset includes both screening and diagnostic exams. Images were captured using equipment from Siemens, IMS, and Planmed. To protect patient privacy, all identifiable information was removed from DICOM metadata and image annotations, with additional masking applied to textual information in the image corners. The pseudonymization process underwent manual validation by human reviewers. The dataset offers both breast-level assessments and lesion-level annotations. The dataset was divided into training (80\%) and testing (20\%) subsets using an iterative stratification algorithm to ensure balanced representation of key attributes, such as BI-RADS categories, breast density levels, and finding types. Examples of VinDr-Mammo images are shown in Figure~\ref{fig:image_samples}. 

For both datasets, pre-processing was performed as specified by the dataset authors \citep{pham2022vindr, sorkhei2021csaw}. For the VMLO-real dataset, an additional image resizing to 512$\times$512 was performed to meet the input requirements of the generative model.

\begin{figure*}[t]
\centering
    \begin{tabular}{cc}
        \includegraphics[width=0.4\textwidth]{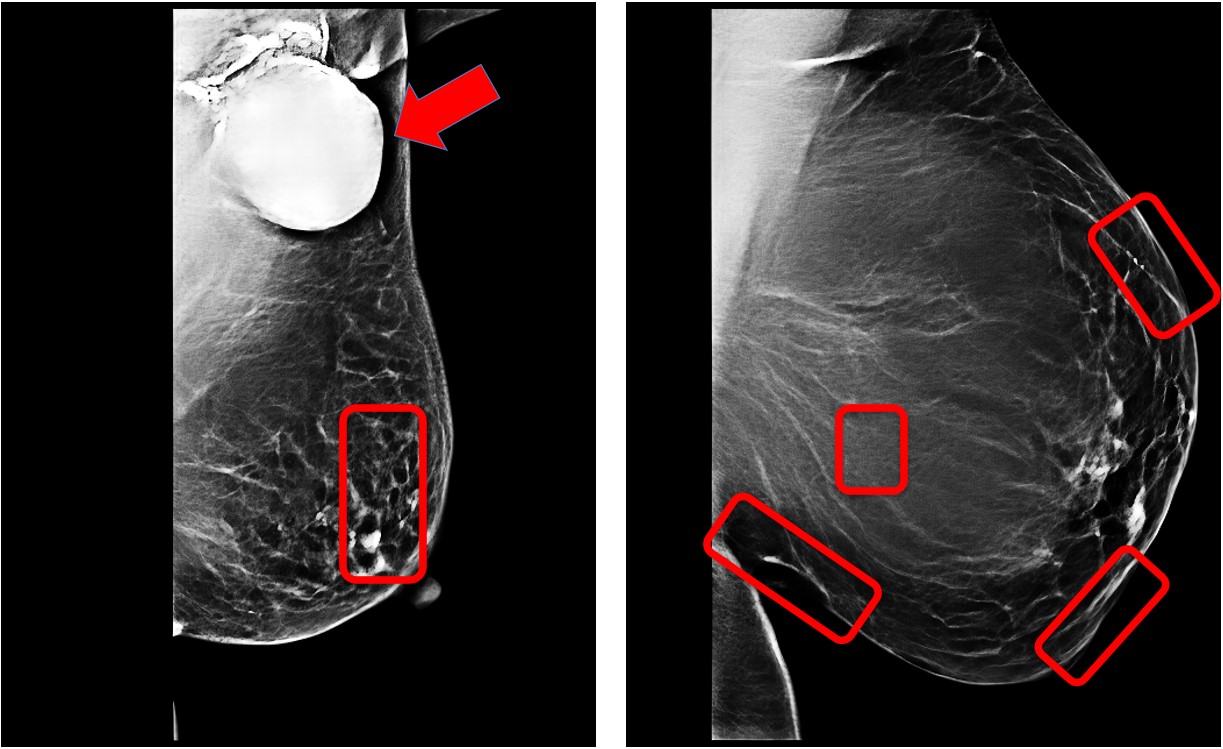} &
        \includegraphics[width=0.4\textwidth]{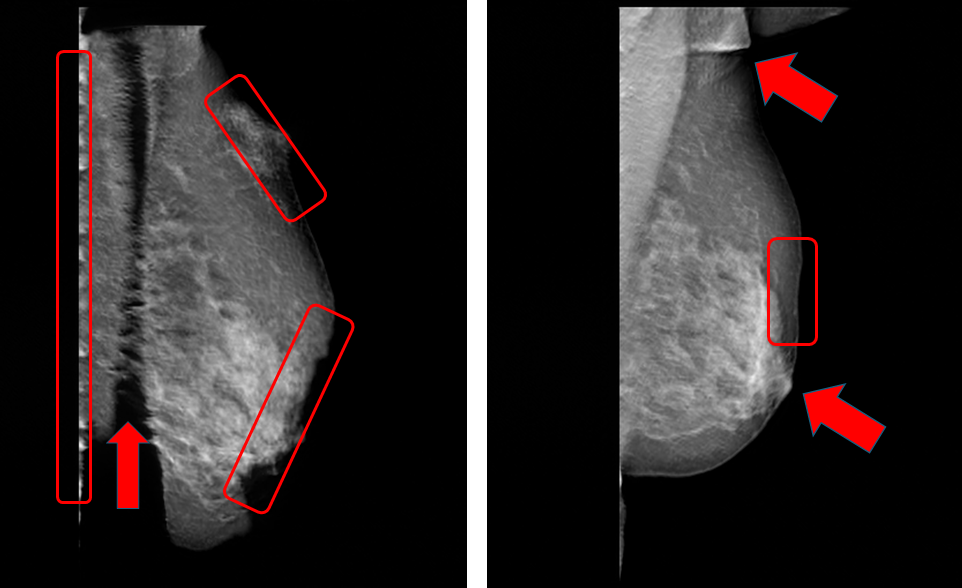} \\
        (a) CSAW-M synthetic & (b) VinDr-Mammo synthetic  \\
    \end{tabular}
\caption{Examples of synthetic mammography images generated from (a) CSAW-M and (b) VinDr-Mammo datasets. Red annotations are areas with unrealistic features that deviate from real patient mammographic anatomy.}
\label{fig:synExamples}
\end{figure*}

The Fr\'echet Inception Distance (FID) scores \citep{heusel2017gans} for the two synthetic datasets are 22 (CSAW-syn) and 43 (VMLO-syn), and the Inception Scores \citep{salimans2016improved} are 2.17 (CSAW-syn) and 2.16 (VMLO-syn), indicating reasonable visual quality overall, even though some images contain artifacts (Figure~\ref{fig:synExamples}). Note that although FID scores may vary based on the generative model and its optimization \citep{lucic2018gans}, in case of synthetic medical images and mammography images, these values have been reported to lie in the range of 10-50 \citep{saragih2024using, oyelade2022generative, rai2024next}. The FID scores of our synthetic datasets also lie within this range. However, these findings illustrate the limitations of dataset-level metrics, such as FID, in reliably detecting shape artifacts in individual images, emphasizing the need for more granular evaluation methods.

\begin{figure}[!h]
\centering
\includegraphics[width=0.8\columnwidth]{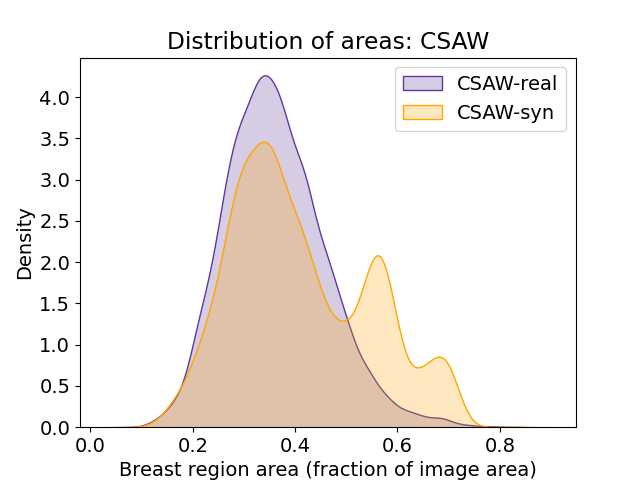}
\\[0.3em]
{\small (a) CSAW}
\\[1em]
\includegraphics[width=0.8\columnwidth]{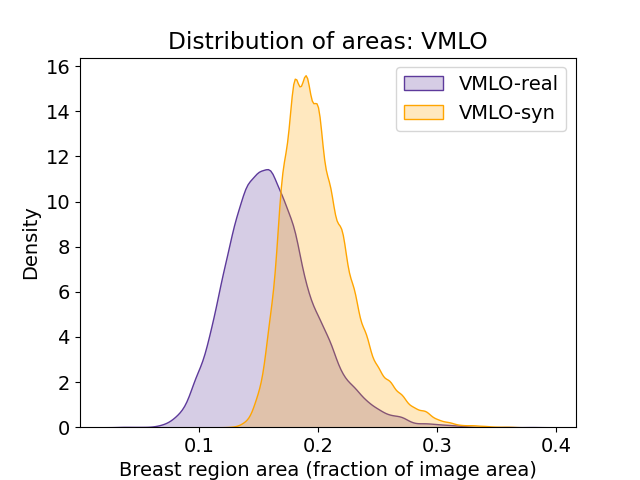}
\\[0.3em]
{\small (b) VMLO}
\caption{Distribution of breast areas as a fraction of the total image area. In case of CSAW-syn, breast area is extrapolated beyond the real distribution whereas in VMLO-syn, the breast area distribution appears shifted towards larger breast areas.}
\label{fig:areas}
\end{figure}

The patient and synthetic data distributions for breast area are shown below (Figure~\ref{fig:areas}). Neither synthetic dataset exactly matches the breast area distribution in the corresponding patient dataset. The CSAW-syn dataset extrapolates beyond the real breast area distribution, and generates breasts of larger sizes than those in the corresponding training dataset. On the other hand, in VMLO-syn, the breast area distribution is shifted (biased) as compared to the corresponding patient dataset. This indicates that in both cases, 1) larger breast shapes than those in the patient dataset were generated and that 2) the prevalence of large breast shapes as seen in the patient dataset was greater than expected in the synthetic dataset. Thus, in both cases, the original distribution of breast area is clearly not maintained.

\subsection{Network-induced Shape Artifact Detection}
Our method (Figure~\ref{fig:overview}) consists of boundary extraction, a novel feature construction and fine-tuning algorithm, and artifact detector. 

\subsubsection{Boundary Extraction and Feature Space Construction}
\label{sec:feature}

The algorithm has two main stages: (1) boundary extraction and (2) feature construction and finetuning.

\textbf{Boundary extraction and tracking} To generate a feature representation of the anatomy of interest, the region of interest (breast in this use case), is segmented via thresholding and morphological operations. Then, a set of boundary pixels, $\mathcal{P} = \{p_{r,c} \in \mathcal{N}^2\}$, is extracted from the segmented mask, where $r,c$ respectively indicate the row and column indices of a boundary pixel $p$; straight edges of the imaging window are excluded from $\mathcal{P}$. Disjointed boundary sections are then connected via morphological opening to ensure robustness when the anatomy of interest exceeds the field of view.

The boundary extraction process involves three steps: image pre-processing, removal of unwanted edges, and ensuring boundary connectivity.
\begin{enumerate}
    \item Image pre-processing consists of the following sub-steps:
    \begin{enumerate}
        \item Gaussian filtering with $\sigma=0.5$ to reduce noise
        \item Thresholding (set to 0) : This threshold may vary based on the dataset. In case of background noise, it may be chosen adaptively or as a percentile of the intensity distribution of an image.
        \item Morphological cleaning: This involved filling holes and morphological opening to obtain a compact object especially in images with great variance in anatomical contrast.
        \item Masking: Obtain the largest connected object, i.e., the image mask. Note that a different segmentation strategy could also be employed instead of the steps above. 
        \item Boundary extraction: Obtain the boundary of the mask by subtracting the mask from its dilated version. This is followed by skeletonizing the boundary. 
    \end{enumerate} 
    \item Removal of straight edges corresponding to the imaging window may not be required for imaging modalities other than mammography, if the anatomy of interest is not bound by the straight edges of an imaging window. It is performed as follows:
    \begin{enumerate}
        \item The top horizontal edge and the vertical side edge are both eliminated by identifying the co-ordinates corresponding to the mode of the boundary co-ordinates in this region.
        \item Any small floating edges are removed.
    \end{enumerate}
    \item The last step is to ensure connectivity of the extracted border. If multiple skeletons are still present, binary morphological closing and skeletonization are recursively performed for increasingly greater breaks in skeletons. Eventually, the largest fully connected skeleton is retained.
\end{enumerate}

\textbf{Feature construction and fine-tuning} The boundary points in $\mathcal{P}$ are ordered by tracking from the top-left and along the anatomical curvature, resulting in a vector $\mathbf{b} = (b_1, b_2, ..., b_K)$. The process starts at the top-left boundary point ($b_1 = \min_{r,c} P$) and follows two rules: (i) $b_{k+1}$ lies within at most the $5\times5$ neighborhood of the current point $b_k$, and (ii) it has the smallest angular gradient relative to $b_k$ and $b_{k-1}$. The tracking terminates when no boundary point is found in the neighborhood, with optional truncation to exclude chest wall regions.

The angular trajectory $\mathbf{a} = (a_1, a_2, ..., a_{K-1})$ is computed from the ordered boundary points as $a_k = \angle(b_k, b_{k+1}) = \textrm{tan}^{-1}\frac{b_{k+1,r} - b_{k,r}}{b_{k+1,c} - b_{k,c}}$. The angular gradient vector $\mathbf{a'}$ is then calculated to capture changes in anatomical shape. To normalize for variations in size, $\mathbf{a'}$ is binned into $N$ equal bins (chosen empirically) and summarized using the cumulative sum (cusum) of angle gradients within each bin. This representation aggregates local shape variations while maintaining global correspondence between bins and anatomical regions, regardless of size differences. Each image, patient or synthetic, is represented as a $N$-dimensional vector in this feature space, which preserves the details of shape for downstream analysis. Note that increasing the bin count captures more local effects, while reducing the bin count emphasizes more global effects. We empirically selected $N=64$ as an optimal compromise between the two factors. However, this parameter can be refined by users based on their prior knowledge of the artifact sizes they aim to capture, providing flexibility to tailor the method to different use cases.

\begin{figure*}[!t]
    \centering
    \includegraphics[width=0.9\linewidth]{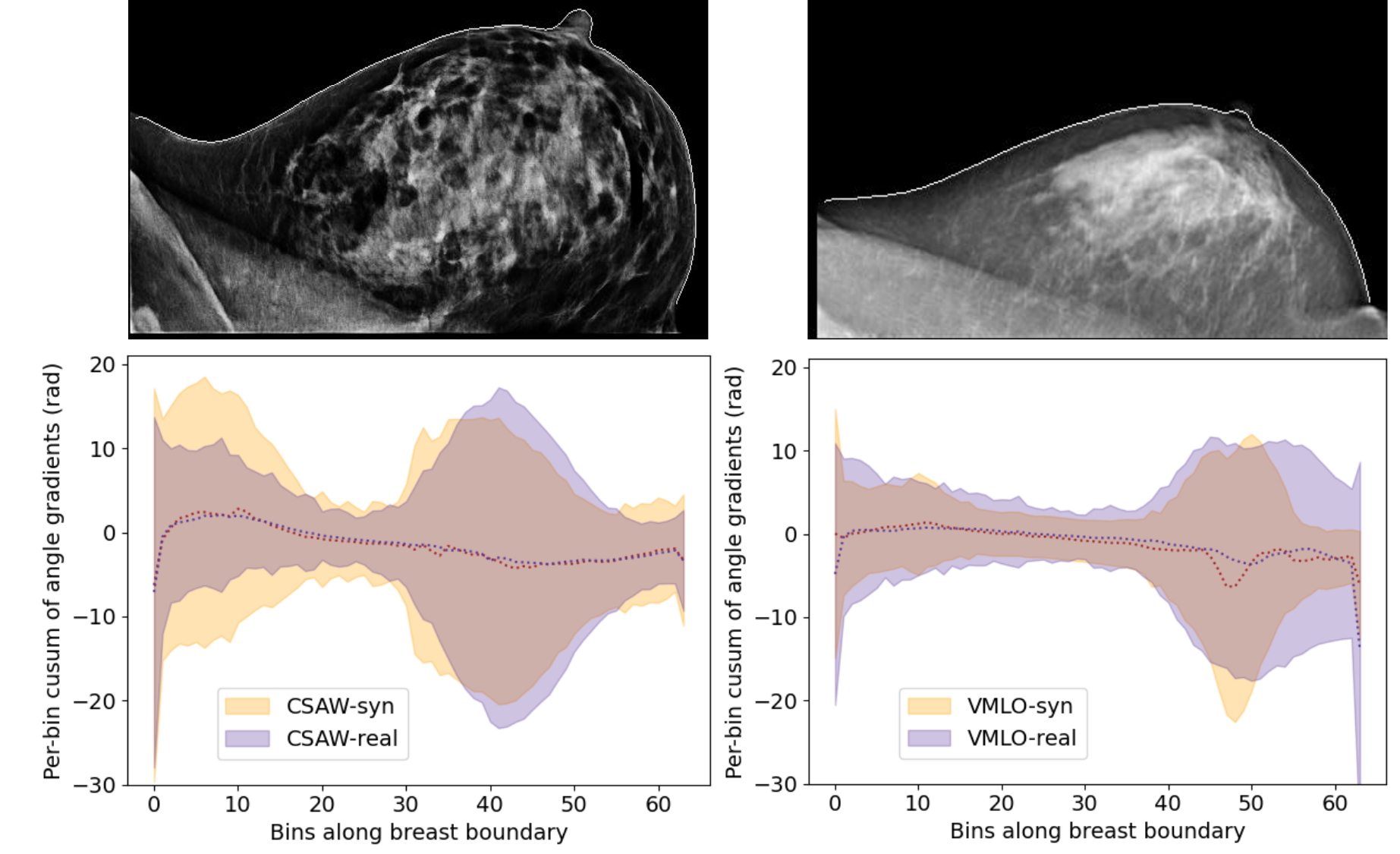}
    \caption{Distributions of bin-wise cumulative sum of angle gradients (dotted line: mean, shaded: one std) show substantial but incomplete overlap between patient and synthetic datasets. Left-to-right bins represent breast shape from top to bottom.}
    \label{fig:distributions}
\end{figure*}

The feature representations from the previous step are summarized per image to capture typical shape variation rates in anatomical shape. For example, the sharp variation associated with the presence of a nipple is expected to occur only once in an image and any greater rate of occurrence might be indicative of an artifact. The 64-d bin-wise representation is then mapped to a 16-dimensional vector (per-image) as follows. The bin edges of the 16-d vector are determined from the approximate range (1-99 percentiles) of cusum values in the \emph{patient} data distribution and the extreme bins are kept open. Each per-image feature vector from the previous step is binned into a 16-d vector accordingly for a given dataset. Extreme bins are eliminated for robustness and the resulting patient and synthetic feature vectors serve as inputs for artifact detection.

Since the length of the breast trajectory is strongly correlated with breast area (Pearson's R = 0.9), and we have observed synthetic data do not necessarily maintain the same distribution of area as the paired patient data, binning and per-image summarization ensures robustness to the effects of varying area.

\subsubsection{Artifact Detection}
For detecting network-induced shape artifacts, we employ an isolation forest (iForest), an established unsupervised anomaly detection algorithm \citep{liu2008isolation}. Although iForest has been used in various applications \citep{al2021isolation}, its application towards identifying network-induced artifacts in synthetic medical images is novel. 

Specifically, from the patient dataset ($X$) as represented in the feature space described in Section~\ref{sec:feature}, a subsample of a dataset ($X'$) is selected. Next, $X'$ is recursively partitioned over a random subset of its features to construct a tree $T$ until each observation is isolated. Several such isolation trees are constructed and together they constitute an isolation forest. The number of trees (100) and the subsampling size (256) are set to the default values \citep{liu2008isolation} but can be adjusted as desired for different applications. The isolation forest yields an anomaly score for each observation based on the average number of partitions required to isolate it across the forest. This score is determined by factors such as the path length to isolate an observation, the average path length over the isolation trees, and the subsampling size. Negative scores indicate outliers, while positive and near-zero scores correspond to inliers. The isolation forest trained on the patient dataset is then employed for prediction on the synthetic dataset. Thus, each synthetic image receives a score, and a rank based on this score.

The code for feature extraction and anomaly detection using the proposed method is available on Github \citep{ShapeCheck}.

%%%%%%%%%%%%%%%%%%%%%%%%%%%%%%%%%%%%%%%%%%%%%%%%%%%%%%%%%%%%%%%%%%%%%%%
% Results
%%%%%%%%%%%%%%%%%%%%%%%%%%%%%%%%%%%%%%%%%%%%%%%%%%%%%%%%%%%%%%%%%%%%%%%

\section{Results}
We present results from the feature extraction process, as well as three sets of results from the proposed shape artifact detection method--dataset-level, image-level, and reader study results.

\subsection{Feature Extraction Results}
 Figure~\ref{fig:distributions} shows the distribution of the bin-wise cusum of angle gradients in synthetic and real datasets. The dotted lines represent the bin-wise mean, while the shaded regions indicate one standard deviation for each bin. The bins along the X-axis correspond to different sections of the breast boundary, arranged sequentially from top to bottom of the breast region. These distributions reflect the typical breast shape for a dataset. Specifically, the early bins correspond to the chest wall, followed by low-variance bins representing the breast region up to the nipple. High standard deviation in the right half indicates the nipple region, while final low-variance bins correspond to the lower breast boundary.

In Figure~\ref{fig:distributions}, we observe that the distributions of the two \emph{patient} datasets are different, despite exhibiting similar trends across bins. The differences may arise from the distinct patient populations in the Swedish (CSAW-real) and Vietnamese (VMLO-real) datasets. Next, the patient and synthetic datasets show substantial, but not perfect overlap for both CSAW-syn and VMLO-syn, suggesting that breast shape is not entirely preserved in the synthetic datasets and that anomalous images may be present. Notably, in the VMLO-syn dataset, the bin-wise means (dotted lines in Figure~\ref{fig:distributions}) are clearly distinct from the corresponding patient dataset, indicating a bias in the synthetic dataset toward a specific breast shape.

\begin{figure}[!b]
\centering
\begin{tabular}{cc}
    \includegraphics[width=0.45\columnwidth]{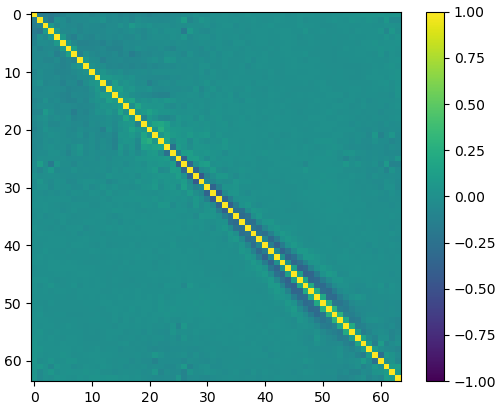} &
    \includegraphics[width=0.45\columnwidth]{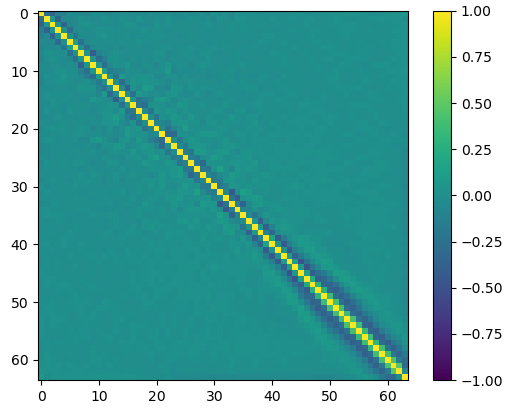} \\
    (a) CSAW-real & (b) VMLO-real  \\
\end{tabular}
\caption{Cross-correlation matrices of patient images in the proposed feature space indicate only moderate correlation among some neighboring feature dimensions, and are largely diagonal otherwise.}
\label{fig:crosscorr}
\end{figure}

\subsubsection{Feature similarity between patient and synthetic datasets}
For the proposed 64-dimensional feature space, the cross-correlation of the feature vectors of the patient datasets are shown in Figure~\ref{fig:crosscorr}. Note that neighboring components demonstrate correlations as they constrain each other according to the possibilities of breast shape. Thus, conventional dimensionality reduction methods (e.g., principal component analysis) do not provide substantial improvement in representational efficiency at this stage.

We quantified the differences between patient and synthetic distributions as follows. Approximately 9,000 samples were selected from both patient and synthetic datasets. For each patient sample, we calculated the Mahalanobis distance relative to the mean and covariance of the patient dataset, creating a distribution of distances. The same procedure was applied to all synthetic samples, using the mean and covariance of the patient dataset. To compare these distributions, we performed a Kolmogorov-Smirnov (KS) test, yielding the following KS-statistics: 0.068 (p-value = 8.27e-19) for CSAW, and 0.13 (p-value = 2.52e-66) for VMLO. This indicates that patient and synthetic distributions have distinct distributions and anomalous images are present in this feature space.

\begin{figure*}[!t]
    \centering
    \includegraphics[width=0.9\linewidth]{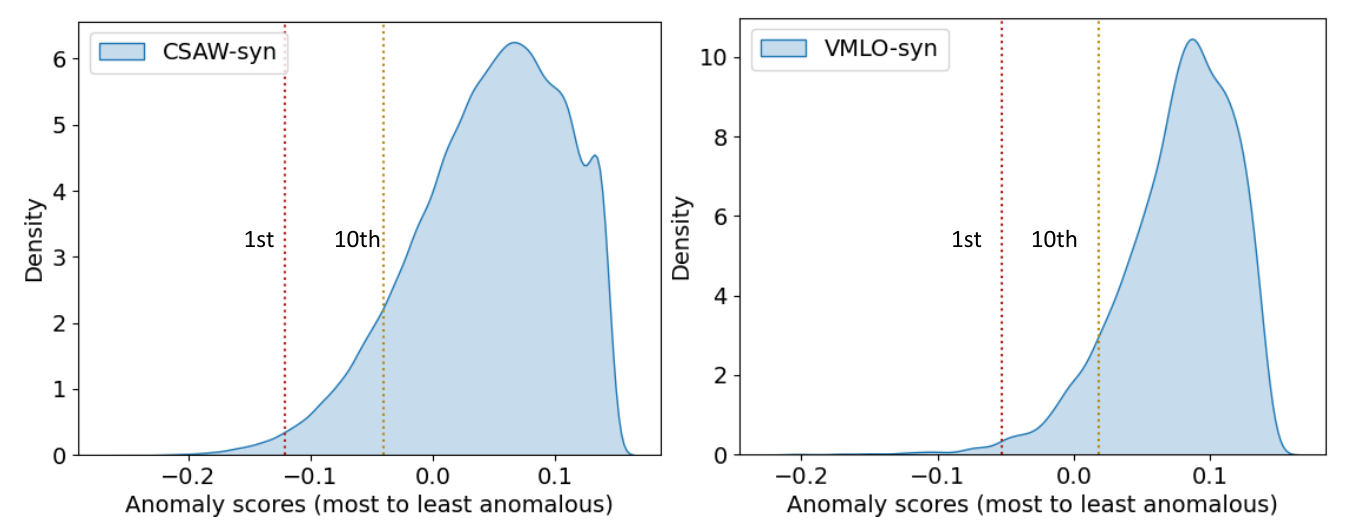}
    \caption{Distribution of anomaly scores for the two synthetic datasets. 1$^{st}$ and 10$^{th}$ percentile are marked.}
    \label{fig:anomaly_scores}
\end{figure*}

\subsection{Dataset-Level Results}
The distributions of anomaly scores for both synthetic datasets are shown in Figure~\ref{fig:anomaly_scores}. Anomaly scores from the isolation forest are bounded between -1 and 1 on the X-axis. Decreasing scores along the negative axis (from 0 to -1) correspond to increasingly anomalous images. In contrast, positive values and values close to zero signify normal images or those with minimal network-induced shape artifacts. Both distributions are left-tailed, suggesting that only a small fraction of the synthetic dataset contains highly anomalous shapes. Thresholds corresponding to the 1$^{st}$ and 10$^{th}$ percentiles of the most anomalous images are marked in Figure~\ref{fig:anomaly_scores}.Three quantile partitions are defined in the distribution of artifact scores, categorized as follows: P1 ( 1$^{st}$ percentile) represents the images with the most obvious or pronounced network-induced shape artifacts, P2 (1$^{st}$–10$^{th}$ percentile) includes images with moderate network-induced shape artifacts, and P3 (10$^{th}$–100$^{th}$ percentile) corresponds to images with minimal or no visible shape artifacts. The categorization of anomaly scores into three partitions is a reasonable starting point when the prevalence of anomalies in a dataset is unknown. These partitions can be further customized and fine-tuned given prior knowledge of the task or the approximate rate of images with artifacts.

To assess the robustness of our proposed method to shift in breast area distributions, each patient-synthetic dataset pair was partitioned according to the quartiles of the breast area distribution in the patient dataset. Distinct partitions in the synthetic dataset were created by matching the partition thresholds determined from the patient dataset. The proposed method was then individually employed on each matched patient-synthetic partition to obtain anomaly scores. An overall anomaly ranking was obtained over the entire dataset based on the anomaly scores from all partitions. It was observed that this global ranking obtained over the area-wise application of the method had a strong correlation (Spearman's $\rho = 0.93$ for both datasets) with the rankings obtained from employing the method on the entire dataset at once. 

\subsection{Image-Level Results}  

\begin{figure*}[!t]
    \centering
    \includegraphics[width=0.9\linewidth]{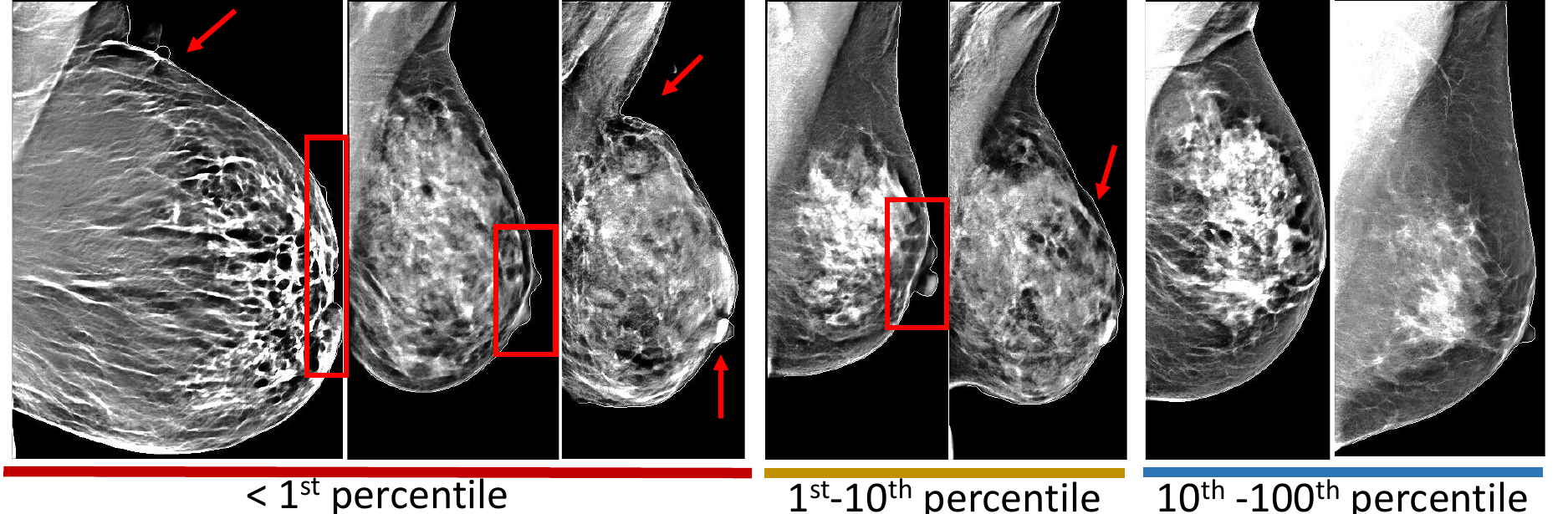}
    \caption{Most to least (L-R) artifact images from CSAW-syn as ranked by our method. Annotations in red are for display only.}
    \label{fig:images_csaw}
\end{figure*}

\begin{figure*}[!t]
    \centering
    \includegraphics[width=0.9\linewidth]{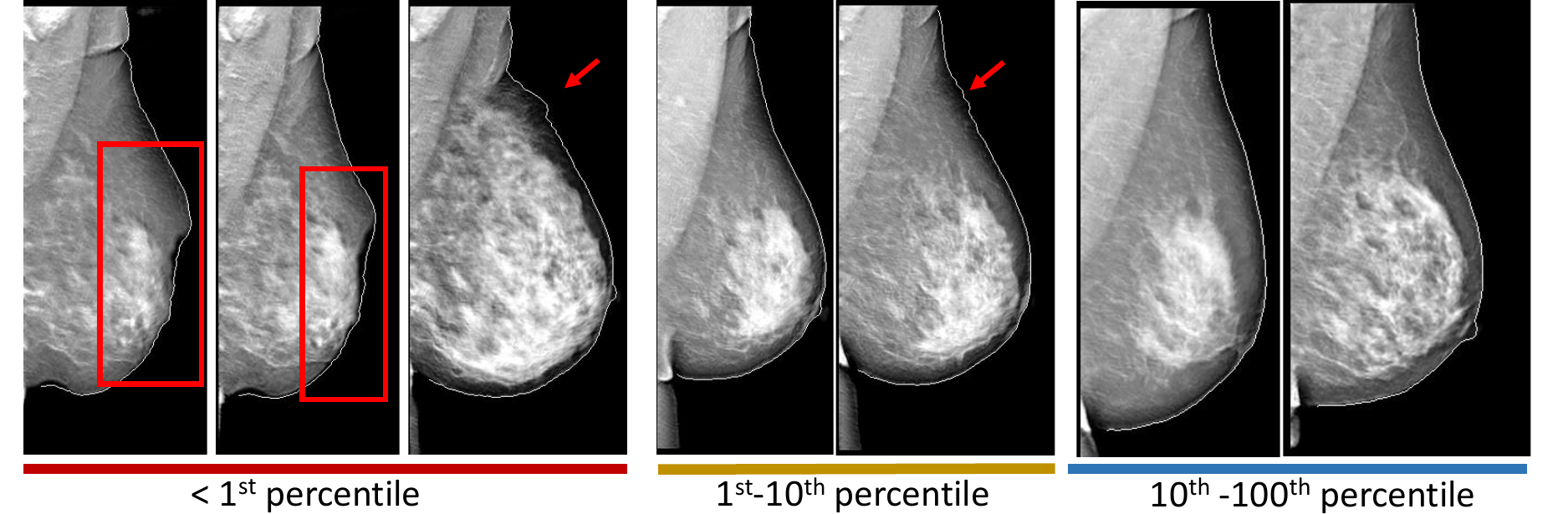}
    \caption{Most to least (L-R) artifact images from VMLO-syn as ranked by our method. Annotations in red are for display only.}
    \label{fig:images_vmlo}
\end{figure*}

Figure~\ref{fig:images_csaw} and Figure~\ref{fig:images_vmlo} present examples of hand-selected images from each partition as ranked by the proposed method for CSAW-syn and VMLO-syn, respectively. As shown in the figures, images in P1 ($\leq 1\%$) have visible shape artifacts. In P2 (1\%-10\%), CSAW-syn shows clear artifacts, while VMLO-syn exhibits minor distortions. In P3 (10\%-100\%), both datasets display well-formed breast shape without visible artifacts.

The figures also highlight that each synthetic dataset exhibits different types of artifacts. In CSAW-syn, artifacts are primarily local, such as multiple nipples, poorly formed nipple regions, and sharp chest wall angles. In contrast, VMLO-syn shows more global artifacts, with visibly malformed breast shape (below the 1st percentile in Figure~\ref{fig:images_vmlo}), alongside minor local artifacts (1st-10th percentile) like non-smooth or angular boundaries. 

\begin{figure*}[!t]
    \centering
    \includegraphics[width=0.9\linewidth]{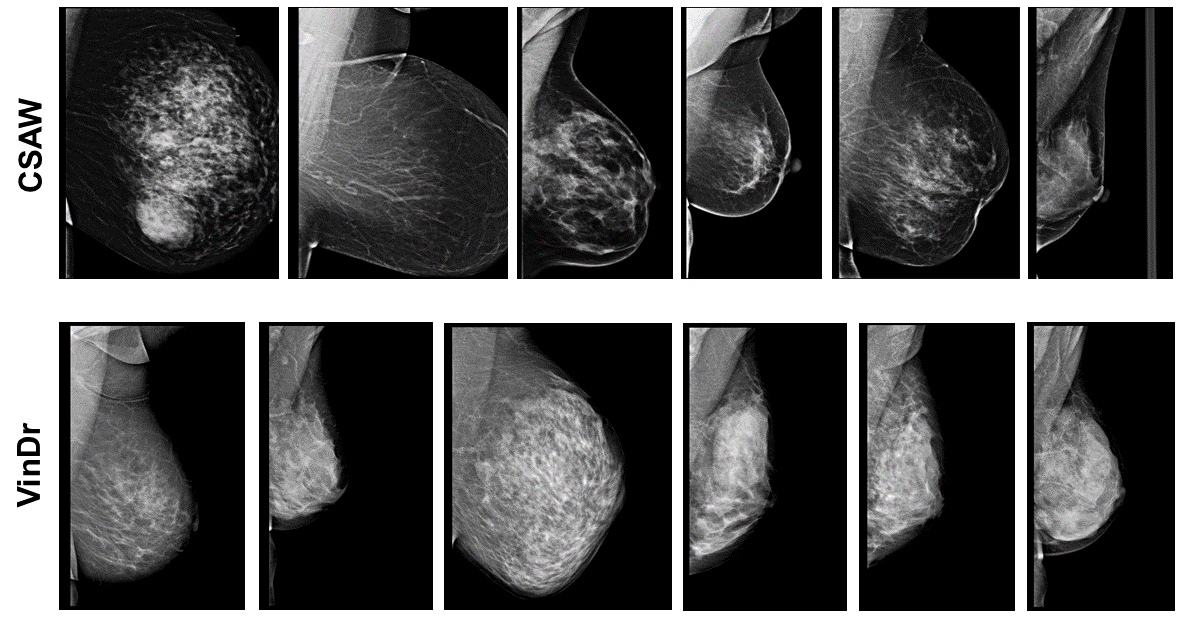}
    \caption{Examples of images identified as anomalous when the proposed method is applied to patient datasets. The first row shows CSAW-real, and second row shows VMLO-real. Practically observed, yet unusual phenomena, such as inadequate tissue coverage, skin folds, missing nipples, etc., can be observed in the identified images.}
    \label{fig:anomalies_patient}
\end{figure*}

The ability of the proposed method to detect artifacts not seen in patient data was explored by using the fitted isolation forests to make predictions on the corresponding patient datasets. The most anomalous images identified within the patient data highlighted cases of imaging issues such as inadequate tissue coverage and missing nipples, as well as atypical breast outlines which can occur due to surgical scars. Examples of these natural shape artifacts are shown in Figure~\ref{fig:anomalies_patient}.

For the synthetic datasets, while some artifacts in P2 resembled the most anomalous images in the patient datasets, artifacts typically observed in P1 were not observed in patient data, and thus, originated from the generative process itself. This confirms that the method can identify artifacts not present in the training dataset. It is important to note that the method can detect these network-induced artifacts without requiring prior knowledge of the anatomy nor any provided labels.

\subsection{Results from the Reader Study}
\begin{figure*}[t]
\centering
\begin{tabular}{cc}
    \includegraphics[width=0.4\textwidth]{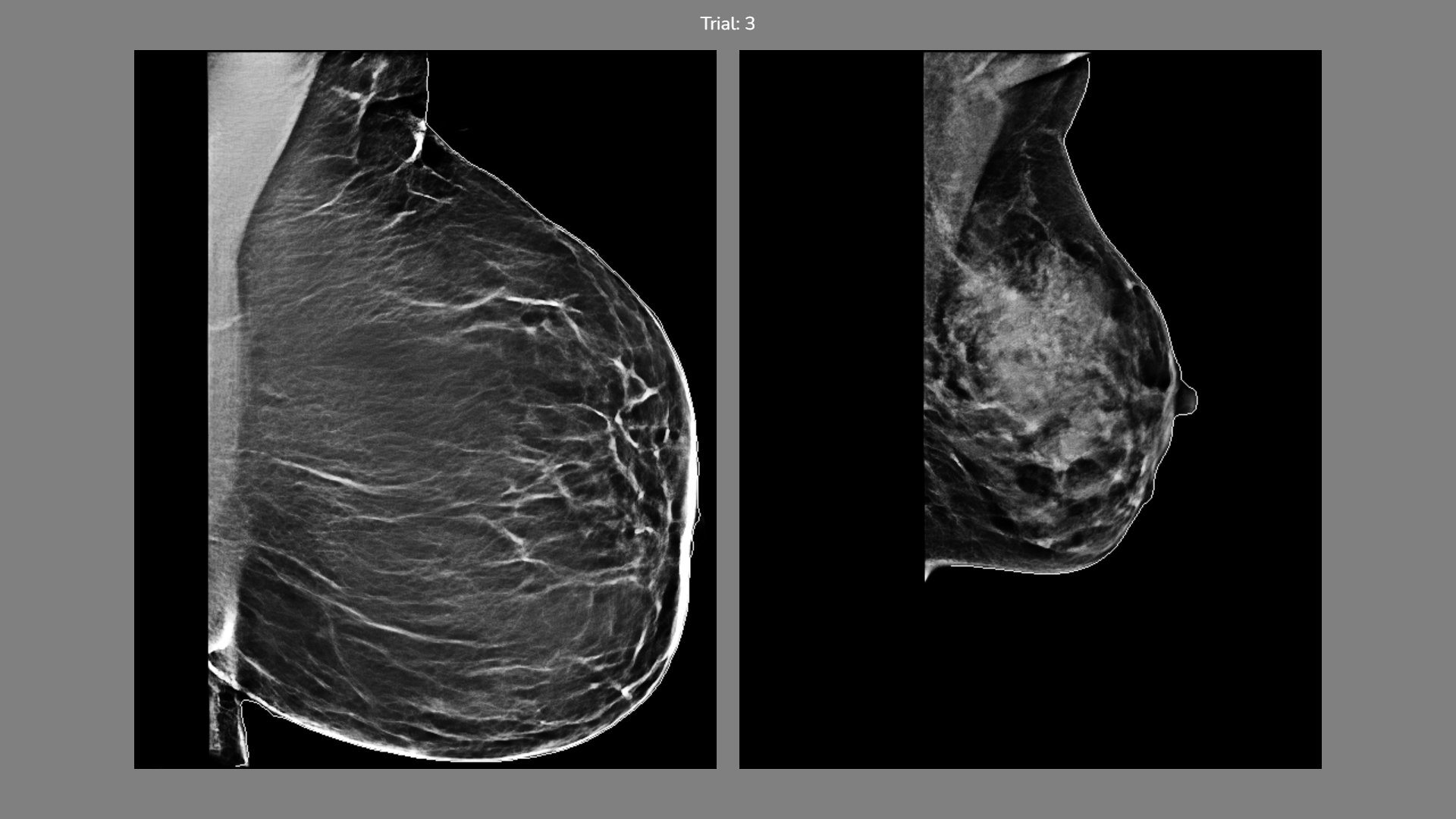} &
    \includegraphics[width=0.4\textwidth]{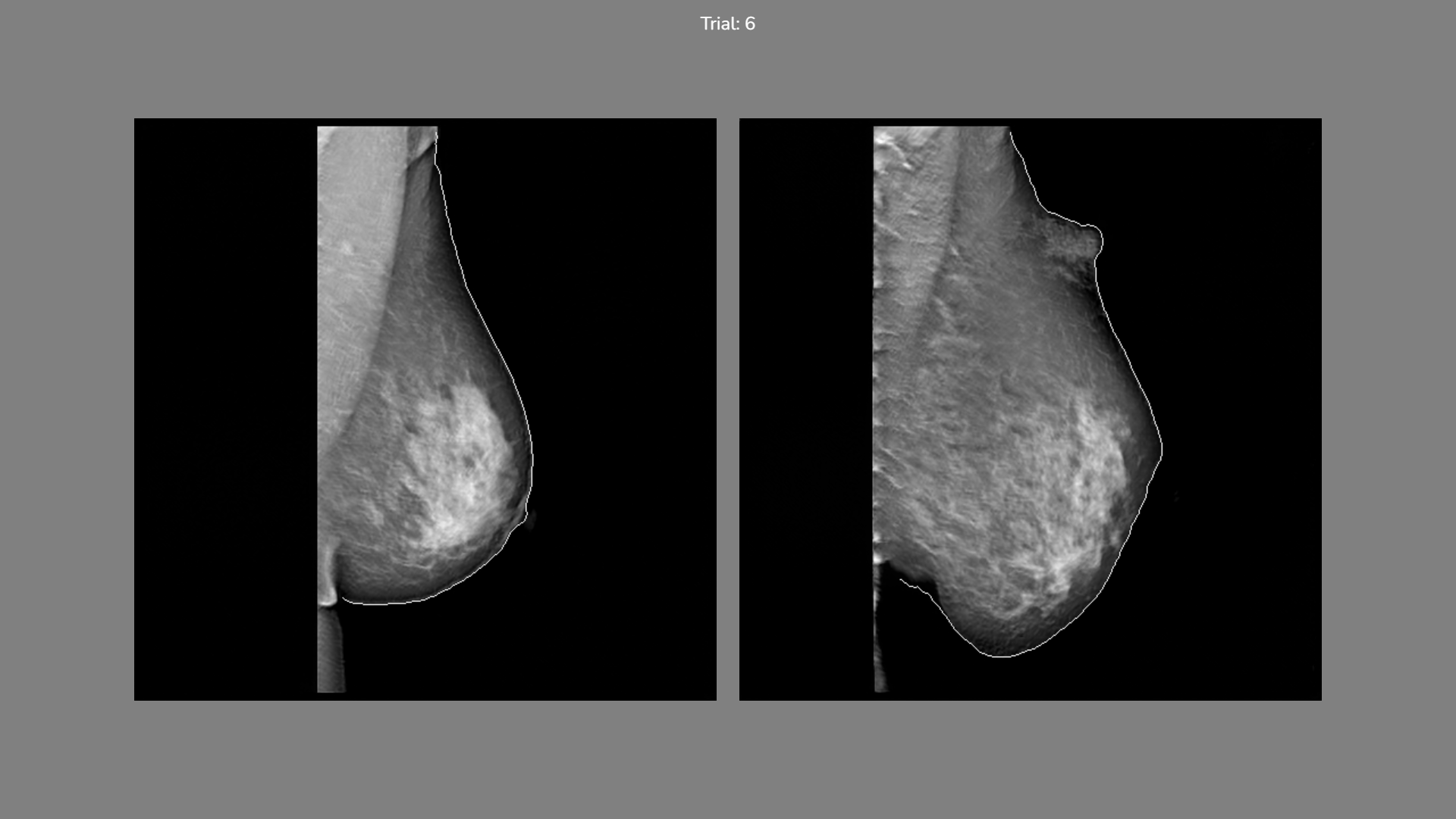} \\
    (a) CSAW-syn dataset pair & (b) VMLO-syn dataset pair  \\
\end{tabular}
\caption{Examples from the reader study conducted for the CSAW-syn and VMLO-syn datasets. Readers were tasked with evaluating pairs of images and selecting the image with more pronounced abnormal shapes. (a) The left panel shows an example from the CSAW-syn dataset with a P1 image on the left and a P1 image on the right. (b) The right panel displays an example from the VMLO-syn dataset with a P2 image on the left and a P1 image on the right.}
\label{fig:reader_study}
\end{figure*}

A 2-Alternative Forced Choice (2-AFC) reader study was conducted separately for each synthetic dataset using the \href{https://gitlab.com/malago/simplephy}{SimplePhy} tool \citep{lago2021simplephy}. Three imaging scientists participated as readers. Readers were presented with pairs of images and asked to select the one they considered to have the most anomalous shape. An example screenshot from the study interface is shown in Figure~\ref{fig:reader_study}. The viewing settings were carefully calibrated for each reader to ensure consistency, and each experiment was completed in a single reading session.

Thirty images from each dataset were selected for the reader study, with ten images drawn from each of the three partitions: $\leq$1\%, 1-10\%, and 10-100\%. This sampling ensures sufficient representation of images that were ranked most anomalous by the algorithm (tail of the anomaly score distribution). All image-pair combinations of the thirty selected images, including image-pairs within the same partition, which resulted in a total of 435 trials, were shown to the readers as illustrated in Figure~\ref{fig:reader_study}. Readers were instructed to select the image with stronger shape artifacts from each side-by-side comparison.

Results are summarized in Table~\ref{tab:reader_study} as the percentage of trials (mean$\pm$std) in which an image was identified as having the most pronounced shape artifacts, reported for each partition. For both datasets, the first partition ($\leq$1\%) was consistently identified as containing the most shape artifacts by all readers, with mean values approximately 1.5-2 times higher than the last partition (10-100\%). This indicates that shape artifacts were effectively concentrated in the first partition, greatly improving search efficiency compared to random visual searches. Further, the similar mean values for the first partition ($\approx$66\%) suggest that all readers consistently found artifacts in this partition to be most distinctive compared to other partitions.

Irrespective of the presence of artifacts, if images lying in the same partition are highly similar, a consistent \emph{intra-partition} ranking cannot be obtained. In our reader study, 10 images are chosen from each partition. Thus, each image is compared with 9 images from the same partition and 29 images over all three partitions. If intra-partition similarity is extremely high, an image is chosen 50\% of the times (equal probability in a 2-AFC) in all its intra-partition trials. This translates to its selection at a rate of about 17\% (1/6) on average over the experiment. This is computed as follows:
\begin{equation}
\text{Image chosen (\%)} = P_{\text{random}} \times \frac{n_{\text{image}}}{N_{\text{total}}}
\end{equation}

where:
\begin{itemize}
    \item \( P_{\text{random}} \): the intra-partition probability of random selection of an image.
    \item \( n_{\text{image}} \): the intra-partition trial count for the specific image.
    \item \( N_{\text{total}} \): the total number of trials conducted in the experiment for the specific image.
\end{itemize}

In our case:
\begin{equation}
\text{Image chosen (\%)} = \frac{1}{2} \times \frac{9}{29}
\end{equation}

Thus, an image will be chosen at least 17\% of the times and not more than $100-17=83\%$ of the times on average if intra-partition similarity is extremely high. The 17\% selection rate will be lower (and the corresponding upper-bound will be higher) if images with varying degrees of artifacts are present in the same partition.

Additionally, mean values decreased monotonically across the three partitions for all readers in both datasets, confirming that successive partitions contained fewer shape artifacts. This is in contrast to the equal mean value ($\approx$50\%) for all partitions expected if images were chosen at random. Further, the second partition is only slightly higher than the third partitions but less so for the VMLO-syn dataset, reflecting a lower fraction of shape artifacts in VMLO-syn compared to CSAW-syn. This is supported by the score distributions shown in (Figure~\ref{fig:anomaly_scores}, where the second partition in CSAW-syn contains negative scores, while that in VMLO-syn straddles zero. 

Kendall-Tau correlations \citep{kendall1938new} between reader rankings and the algorithm rankings over all images in the reader study were 0.45 for CSAW-syn (reader values: 0.43, 0.51, 0.40) and 0.43 for VMLO-syn (reader values: 0.33, 0.42, 0.55), indicating reasonable agreement between the two. Note that high Kendall-Tau values are \emph{not} expected due to low visual distinguishability between images with close rankings.  

Finally, the AUC (area under the curve) values were computed according to \citep{liu2008isolation, hand2001simple} based on the anomaly rankings within the set of the images employed in the reader study, where true anomalies were defined based on the mean values from the reader study. The resulting AUC was 0.97 (CSAW) and 0.91 (VMLO), where true anomalies were defined based on reader consensus.

%%%%%%%%%%%%%%%%%%%%%%%%%%%%%%%%%%%%%%%%%%%%%%%%%%%%%%%%%%%%%%%%%%%%%%%
% Discussion and Conclusion
%%%%%%%%%%%%%%%%%%%%%%%%%%%%%%%%%%%%%%%%%%%%%%%%%%%%%%%%%%%%%%%%%%%%%%%

\section{Discussion and Conclusion}
We propose a knowledge-based method to detect network-induced shape artifacts, and demonstrate its use on synthetic mammograms. While similar knowledge-based approaches for characterizing shapes have been used in tasks such as kidney stone classification \citep{duan2013differentiation} and nipple localization in mammograms \citep{zhou2004computerized}, this is the first application of such an approach to synthetic data for detecting network-induced artifacts. Breast shape is associated with breast density, architectural distortions, and demographic features \citep{gaur2013architectural,del2007mammographic}, and unrealistic breast shape may negatively affect downstream task performance. In addition, our method is shape and modality agnostic, and hence highly versatile. 

Although research on characterizing breast tissue is extensive \citep{gastounioti2016beyond}, methods for evaluating breast shape fidelity are limited. Most shape variation in mammography occurs along the breast curvature, while the straight edges of the imaging window show minimal variation. Conventional shape features like compactness or area often miss anatomical inaccuracies along the curvature, making shape artifacts in synthetic mammograms difficult to detect. Our proposed method addresses these challenges effectively.

\begin{table*}[t]
\centering
\scriptsize 
  \caption{Results from the reader study demonstrate that the images ranked worst (P1) by our method were also chosen as the worst by all readers. Highest values in bold.}%
  {\begin{tabular*}{0.7\textwidth}{@{\extracolsep{\fill}}|c|ccc|ccc|}
      \hline
      \bfseries Dataset &  & \bfseries CSAW-syn & & & \bfseries VMLO-syn & \\
      \hline
      \bfseries Reader/Partition & \bfseries P1 & \bfseries P2 & \bfseries P3 & \bfseries P1 & \bfseries P2 & \bfseries P3 \\
        \hline
      Reader 1 & \textbf{65$\pm$9}\% & 48$\pm$12\% & 37$\pm$9\% & \textbf{63$\pm$9}\% & 45$\pm$9\% & 41$\pm$12\% \\
      
      Reader 2 & \textbf{68$\pm$7}\% & 47$\pm$9\% & 34$\pm$8\% & \textbf{67$\pm$9}\% & 42$\pm$11\% & 41$\pm$17\% \\
      
      Reader 3 & \textbf{64$\pm$7}\% & 47$\pm$13\% & 39$\pm$7\% & \textbf{73$\pm$8}\% & 42$\pm$8\% & 34$\pm$14\% \\
      \hline
      Mean of means & \textbf{66}\% & 47\% & 37\% & \textbf{68}\% & 43\% & 39\%\\
      \hline
  \end{tabular*}}
  \label{tab:reader_study}
\end{table*}

Our approach is designed to be widely applicable to synthetic datasets for detecting network-induced shape artifacts, even in the absence of artifact labels. It provides rankings instead of binary decisions on artifact presence, which can improve the efficiency of visual search for artifacts, reducing the burden on expert readers. These rankings can effectively separate images with artifacts from normal images and users can adjust the threshold post-hoc according to their specific requirements, enabling flexible and tailored detection.

Consider a synthetic dataset of 10,000 images of which 5\% contain artifacts. A random sample of 100 images would be expected to yield 5 images with artifacts on average. That is, the rate of artifact discovery is 1 in 20 (= 0.05). Practically, the small sample size (100) may result in even fewer, or no artifacts being discovered. However, if the proposed method is employed with an accuracy of 70\% (highest mean values for both datasets were about 67\%), the same sample would contain about 70 images with artifacts. The new rate of artifact discovery would be 7 in 10 images (= 0.7). Thus, the rate of artifact discovery would be improved 14 times (new rate/ old rate = 0.7/0.05) over random spot checking. More generally, the rate improves as a factor of $\textrm{accuracy (\%)}/\textrm{artifact prevalence (\%)}$, and may differ as the two factors vary.

We hope that our method will assist annotators and domain experts by providing the first step in obtaining labels for a dataset, which can then be used to develop semi-supervised or supervised anomaly detection methods, or to clean synthetic datasets before using them for training or testing AI models.

While we provide an example with breast imaging, the proposed method can be generalized to other anatomies and imaging modalities where the region of interest can be segmented, such as lungs, abdomen, brain, or any other anatomical region of interest. This broad applicability is due to the reliance of the method on the characteristic shape of the anatomy. It is particularly valuable in scenarios where anatomy cannot be described adequately by area-based features or conventional compactness and convexity features.

In lung imaging via chest radiographs, for example, the method can be applied to detect artifacts in the shape and contour of the lungs from synthetic data. By analyzing the boundary of the lung, we can extract the angle gradients along the lung border. These gradients capture the curvature and shape of the lung, which are necessary for detecting any anomalies caused by synthetic artifacts. In this context, the method can identify unusual shapes that do not align with the expected lung morphology, such as irregularities in lung shape or distortions in the pleural surface, which may indicate network-induced artifacts.

Similarly, in abdominal computed tomography (CT), the method can be applied to detect synthetic artifacts in shape for various organs such as the liver and the kidneys. By analyzing the contours of these organs and calculating the angle gradients along their boundaries, we can detect unnatural shapes, such as deformed or overly smoothed organ outlines, which might result from flaws in the synthetic data generation process. The method could flag these artifacts by identifying discrepancies in the natural shape of abdominal organs. Another example could be characterizing brain hemorrhages in computed tomography, which have characteristic shapes based on type, and assessing synthetic tissue lesions, which may have characteristic shapes based on the presence of malignancy. The proposed method can effectively capture such shape variations specific to anatomical structures. In the future, we plan to extend our work to synthetic datasets from other medical imaging modalities in a domain-relevant manner.

A limitation of this work is that the method relies on a continuous boundary from prior segmentation, and we did not analyze the impact of different segmentation methods. In low-quality synthetic datasets, network-induced background artifacts may negatively impact segmentation methods. While this was not observed in our work, it may be relevant for other synthetic datasets. To identify shape artifacts when constant thresholding may be unreliable for segmentation due to background artifacts in synthetic data, an adaptive threshold can be selected for segmentation. This threshold could be chosen as a percentile from the pixel intensity distribution in an image, in a user-informed manner, to ensure consistency with the patient data. A further consideration is the size of shape artifacts. While our method is agnostic to artifact size, prior knowledge about artifact scale or type can be incorporated into the algorithm through bin counts or relative bin locations. Another limitation of our work is that the clinical impact of the identified artifacts was not studied. In the future, a reader study conducted by radiologists on artifacts in synthetic data as well as natural shape artifacts (refer Figure~\ref{fig:anomalies_patient}) could provide valuable clinical insights. Additionally, the present work focuses on breast shape rather than breast tissue. We are currently developing a complementary method for assessing artifacts in breast tissue, which will also require validation by clinical experts. Other future research directions include further exploring the localization of detected artifacts within flagged images, as well as identifying the origin of shape artifacts in relation to the dataset, the model and its optimization.

In conclusion, we have developed a knowledge-based method that efficiently identifies network-induced shape artifacts in synthetic medical images. Being agnostic to the generative process, it is applicable to emerging model architectures and scenarios where artifact characteristics are unknown, making it valuable for quality assurance in large-scale synthetic dataset deployment where visual inspection of individual images is impractical.

%%%%%%%%%%%%%%%%%%%%%%%%%%%%%%%%%%%%%%%%%%%%%%%%%%%%%%%%%%%%%%%%%%%%%%%
% Mandatory Sections. Please complete, especially for final publication
%%%%%%%%%%%%%%%%%%%%%%%%%%%%%%%%%%%%%%%%%%%%%%%%%%%%%%%%%%%%%%%%%%%%%%%

% Acknowledgements.
% Please include any funding, intellectual contributions not included in the authorship, and any other acknowledgements.
\acks{Rucha Deshpande and Tahsin Rahman acknowledge funding by appointment to the Research Participation Program at the Center for Devices and Radiological Health administered by the Oak Ridge Institute for Science and Education through an interagency agreement between the US Department of Energy and the US Food and Drug Administration.}

% Ethical Standards.
% Please edit with the appropriate ethics considerations for your work. Include any pertinent IRB information, etc.
%
% Please note that the submission requirements included:
% The work presented must follow appropriate ethical standards in conducting research and writing the manuscript, following all applicable laws and regulations regarding treatment of animals or human subjects.
\ethics{The work follows appropriate ethical standards in conducting research and writing the manuscript, following all applicable laws and regulations regarding treatment of animals or human subjects.}

% Conflict of Interest
% Declaration of possible conflicts of interest: Authors must disclose any financial, organisational, commercial or personal conflicts of interest that might bias their work.
% If no conflicts, please say "We declare we don't have conflicts of interest."
\coi{We have no conflicts of interest to declare. The mention of commercial products, their sources, or their use in connection with material reported herein is not to be construed as either an actual or implied endorsement of such products by the Department of Health and Human Services. This is a contribution of the U.S. Food and Drug Administration and is not subject to copyright.}

% Data availability
\data{Both patient datasets used in this work are public datasets. One synthetic dataset (SinKove) is also public, while another was generated in-house by training on a real dataset (VinDr).}

\bibliography{melba-bibliography}

\end{document}